\documentclass[letterpaper, 10 pt, conference]{ieeeconf}  
\usepackage{amssymb}
\usepackage{amsmath} 
\usepackage{graphicx}
\usepackage{hyperref}
\usepackage{wrapfig}
\usepackage{booktabs}
\usepackage{multirow}
\usepackage{tabularx}

\usepackage[capitalise,nameinlink]{cleveref}
  \crefname{section}{Sec.}{Sec.}
  \Crefname{section}{Section}{Sections}
  \crefname{figure}{Fig.}{Fig.}
  \Crefname{figure}{Figure}{Figures}
  \crefname{table}{Tab.}{Tab.}
  \Crefname{table}{Table}{Tables}

\IEEEoverridecommandlockouts                           

\title{\LARGE \bf
A Soft Wrist with Anisotropic and Selectable Stiffness
\\for Robust Robot Learning in Contact-rich Manipulation
}

\author{Steven Oh$^{1*}$, Tomoya Takahashi$^{1*}$, Cristian C. Beltran-Hernandez$^{1}$, Yuki Kuroda$^{1}$, and Masashi Hamaya$^{1}$
\thanks{*Equal contributions}
\thanks{$^{1}$ OMRON SINIC X Corp., Bunkyo-ku, Tokyo, 113-0033, Japan.
        {\tt\small masashi.hamaya@sinicx.com}}%
}

\begin{document}

\maketitle
\thispagestyle{empty}
\pagestyle{empty}

\begin{abstract}
    Contact-rich manipulation tasks in unstructured environments pose significant robustness challenges for robot learning, where unexpected collisions can cause damage and hinder policy acquisition. Existing soft end-effectors face fundamental limitations: they either provide a limited deformation range, lack directional stiffness control, or require complex actuation systems that compromise practicality. This study introduces CLAW (Compliant Leaf-spring Anisotropic soft Wrist), a novel soft wrist mechanism that addresses these limitations through a simple yet effective design using two orthogonal leaf springs and rotary joints with a locking mechanism.
    CLAW provides large 6-degree-of-freedom deformation (40mm lateral, 20mm vertical), anisotropic stiffness that is tunable across three distinct modes, while maintaining lightweight construction (330g) at low cost (${\sim}$\$550). Experimental evaluations using imitation learning demonstrate that CLAW achieves 76\% success rate in benchmark peg-insertion tasks, outperforming both the Fin Ray gripper (43\%) and rigid gripper alternatives (36\%). CLAW successfully handles diverse contact-rich scenarios, including precision assembly with tight tolerances and delicate object manipulation, demonstrating its potential to enable robust robot learning in contact-rich domains. 
    Project page: \url{https://project-page-manager.github.io/CLAW/}
\end{abstract}



\section{Introduction}

Autonomous manipulation in real-world environments represents one of robotics' most significant challenges, particularly when robots must perform contact-rich tasks such as assembly, insertion, and surface interaction. 
The increasing deployment of robots beyond structured factory settings into unstructured environments, such as homes, laboratories, and collaborative workspaces, has intensified the need for manipulation systems that can robustly handle unexpected contacts due to environmental variations and uncertainties.

Recent learning approaches, including imitation learning~\cite{zhao2023learning,chi2023diffusion} and reinforcement learning~\cite{nguyen2019review}, show considerable promise for enabling robots to acquire complex manipulation skills in environmental variations. However, these paradigms face challenges of data collection and policy execution in uncertain scenarios. During learning, traditional rigid robots, while precise, can pose substantial risks; for example, partially trained policies or human demonstrators may issue control commands that lead to hazardous collisions, potentially damaging both the robot and its surroundings.

\begin{figure}[t]
    \centering
    \includegraphics[width=\linewidth]{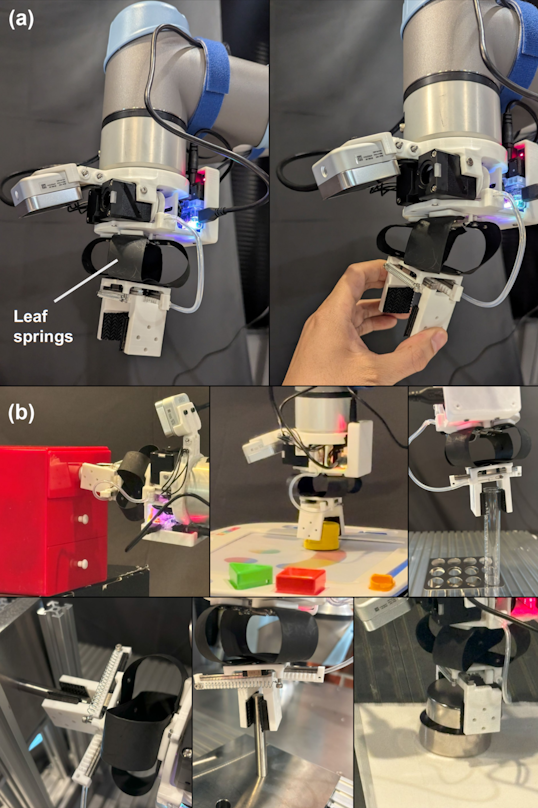}
    \caption{\textbf{The CLAW soft wrist:} (a) A novel soft wrist featuring two orthogonal leaf springs to achieve anisotropic stiffness and enabling 6-DoF compliance. (b) CLAW’s extensive evaluation across a range of challenging contact-rich manipulation tasks using imitation learning.}
    \label{fig:overview}
\end{figure}

Existing approaches to address this challenge fall into two main categories. Force-controlled robots~\cite{haddadin2022franka} provide one solution by enabling compliant interaction, but require sophisticated control algorithms and expensive hardware that limit their accessibility. Alternatively, physically soft end-effectors can be integrated with standard position-controlled robots to provide passive safety through mechanical compliance~\cite{jeong2024bariflex, azulay2025visuotactile, zhao2025learning}. Despite significant advances in soft robotics, existing soft end-effectors struggle with satisfying the following requirements:

    

    \paragraph  {\bf Large and multi-directional deformation} Contact-rich tasks inevitably involve unexpected collisions. To absorb such impacts, the end-effector must be able to deform in 6-degree-of-freedom (6-DoF), and a larger deformation range enables it to accommodate greater uncertainties.
    \paragraph {\bf Anisotropic and adjustable stiffness} Direction-dependent stiffness is desirable. For instance, low vertical stiffness facilitates gentle surface contact, whereas moderate horizontal stiffness both absorbs lateral collisions and prevents the end-effector from sagging when oriented horizontally. Also, adjustable stiffness can broaden the range of tasks.
    \paragraph{\bf Lightweight and durable construction} Reducing mass not only increases the robot’s payload capacity but also diminishes oscillations during abrupt accelerations and decelerations. Moreover, lowering the end-effector’s weight reduces gravitational sag. Because training involves many trials, the end-effector should be robust and durable.
    \paragraph {\bf Low cost and simple implementation} To encourage widespread adoption, the design should minimize part count, rely on readily available components, and be inexpensive and straightforward to assemble.

We propose the novel soft end-effector CLAW ({\bf C}ompliant {\bf L}eaf-spring {\bf A}nisotropic soft {\bf W}rist) for robust robot learning, shown in~\cref{fig:overview} (a). CLAW features a two-finger gripper and a soft wrist composed of two orthogonal leaf springs and rotary joints with a locking mechanism. CLAW exhibits large 6-DoF deformations and anisotropic stiffness and can switch between three stiffness modes.
We also propose using a convex tape for the leaf spring, resulting in a stable, compact wrist structure.
The total construction cost is approximately \$550, and it only weighs 330g.  

The main contributions of this study are the design of a novel soft wrist and its evaluation in learning contact-rich tasks. We compared CLAW with widely-used baselines, a TPU-printed soft Fin Ray gripper, and an off-the-shelf rigid parallel gripper.
In the deformation evaluation, CLAW demonstrated higher compliance in a wider range of directions than the baselines. 
Then, we evaluated CLAW within an imitation learning framework, leveraging the sample-efficient method Action Chunking with Transformer (ACT)~\cite{zhao2023learning} to train manipulation policies on benchmark peg-insertion tasks. CLAW outperformed the baseline grippers. Beyond benchmarking, CLAW demonstrated robust performance across a variety of challenging contact-rich tasks, including delicate object manipulation and tight-tolerance insertions, showcasing its potential to enable robot learning in unstructured environments, shown in \cref{fig:overview} (b).

\section{Related Work}
\label{sec:RelatedWork}
A wide variety of mechanisms have been proposed to realize compliance in robotic manipulation. In this section, we review prior work on soft grippers and soft wrists for contact-rich manipulation.

Numerous soft grippers composed of flexible materials have been developed. One approach places springs along each principal axis to impart tunable, potentially anisotropic stiffness to the hand~\cite{fukaya2024four}, although this inevitably increases part count and mechanical complexity. Tendon-driven, underactuated rigid fingers~\cite{azulay2025visuotactile,morgan2021vision} and hands~\cite{ko2020tendon} have demonstrated robust and dexterous manipulation with relatively few actuators, but their deformation directions remain constrained.
Fin Ray grippers, inspired by the deformation mechanics of fish fins~\cite{crooks2016fin}, have been employed in insertion tasks~\cite{nie2018adaptive,hartisch2023high} and pick-and-place operations~\cite{wang2024fin}. When coupled with a backdrivable motor, they enable adaptive grasping and other contact-rich actions~\cite{jeong2024bariflex}. Despite the advantages of rapid 3D printing and the ability to tune stiffness via geometry and material choice, Fin Ray grippers remain limited in both their range and directions of deformation. Pneumatic grippers and pneumatic tactile sensors~\cite{brahmbhatt2023zero,oller2023manipulation} allow large, multi-directional deformations, but their reliance on external pneumatic hardware poses a practical constraint. Similarly, tendon-driven silicone fingers can perform delicate, contact-rich tasks such as hair grooming~\cite{yoo2025soft}, but they are less suitable for high-force interactions.

\begin{table}[]
\caption{Mechanical characterization of CLAW and other soft wrists}
\label{tab:mech_char}
\begin{tabular}{|cc|ccc|c|}
\hline
\multicolumn{2}{|c|}{\multirow{2}{*}{Specification}} & \multicolumn{3}{c|}{Value} & \multirow{2}{*}{Unit} \\ \cline{3-5}
\multicolumn{2}{|c|}{} & \multicolumn{1}{c|}{\begin{tabular}[c]{@{}c@{}}{\bf Ours}\\ 6-DoF\end{tabular}} & \multicolumn{1}{c|}{\begin{tabular}[c]{@{}c@{}}Wrist~\cite{von2020compact}\\ 6-DoF\end{tabular}} & \begin{tabular}[c]{@{}c@{}}Wrist~\cite{10745776}\\ 2-DoF\end{tabular} &  \\ \hline
\multicolumn{2}{|c|}{Weight} & \multicolumn{1}{c|}{\begin{tabular}[c]{@{}c@{}}330\\ (w/ gripper,\\ wrist, camera)\end{tabular}} & \multicolumn{1}{c|}{\begin{tabular}[c]{@{}c@{}}Not\\ reported\end{tabular}} & \begin{tabular}[c]{@{}c@{}}110\\ (w/o gripper)\end{tabular} & g \\ \hline
\multicolumn{2}{|c|}{Size} & \multicolumn{1}{c|}{\begin{tabular}[c]{@{}c@{}}155 × 150 × 85\\ (W × H × L)\end{tabular}} & \multicolumn{1}{c|}{$\phi$ 75 × 60} & $\phi$ 63 × 45 & mm \\ \hline
\multicolumn{1}{|c|}{\multirow{4}{*}{\rotatebox[origin=r]{90}{Deformation~~~}}} & X/Y & \multicolumn{1}{c|}{40} & \multicolumn{1}{c|}{6.35} & N/A & mm \\ \cline{2-6} 
\multicolumn{1}{|c|}{} & Z & \multicolumn{1}{c|}{\begin{tabular}[c]{@{}c@{}}20 (comp.), \\ 10 (ext.)\end{tabular}} & \multicolumn{1}{c|}{11} & N/A & mm \\ \cline{2-6} 
\multicolumn{1}{|c|}{} & \begin{tabular}[c]{@{}c@{}}Roll/\\ Pitch\end{tabular} & \multicolumn{1}{c|}{15} & \multicolumn{1}{c|}{8} & 20 & Deg \\ \cline{2-6} 
\multicolumn{1}{|c|}{} & Yaw & \multicolumn{1}{c|}{\begin{tabular}[c]{@{}c@{}}30 (Full/Half-lock), \\ 45 (Free)\end{tabular}} & \multicolumn{1}{c|}{\begin{tabular}[c]{@{}c@{}}Not\\ reported\end{tabular}} & N/A & Deg \\ \hline
\multicolumn{2}{|c|}{\begin{tabular}[c]{@{}c@{}}Maximum\\ grip force\end{tabular}} & \multicolumn{1}{c|}{7.5} & \multicolumn{1}{c|}{N/A} & N/A & N \\ \hline
\end{tabular}
\end{table}

Given the limitations of soft grippers, researchers have also explored soft wrists as an alternative approach of achieving compliance, enabling large deformations while working in conjunction with a rigid gripper to deliver sufficient grasping force. As early as the 1980s, the Remote Center Compliance (RCC) device was introduced as a spring-based mechanism to passively compensate for positional errors in assembly~\cite{whitney1982quasi}. However, RCC offers only a limited range of motion~\cite{von2020compact}. Soft wrists~\cite{von2020compact, zhang2023compliant} enable large deformations in 6-DoF, but they are prone to sagging under gravity when the manipulator is oriented horizontally. Counteracting this sag typically requires stiffer or additional springs, which reduces compliance or increases structural complexity.
More recent designs seek variable stiffness. A sheet-type flexure combined with tendon actuation enables a wrist with adjustable stiffness and has demonstrated diverse contact-rich tasks~\cite{10745776}. Nevertheless, this design requires complex structures, including three servomotors and tendons to regulate stiffness. A 3D-printed wrist incorporating a honeycomb lattice has also been proposed to achieve bimodal stiffness responses~\cite{jeong2025biflex}. Although straightforward to fabricate, its deformation range remains narrow. \cref{tab:mech_char} introduces the mechanical characterization between CLAW and the other wrists~\cite{von2020compact,10745776}.

Overall, while they show significant potential, none fully resolve the challenges outlined in the introduction. Existing grippers and wrists require mechanically complex structures, restrict the range or direction of deformation, or struggle to maintain stable grasping in contact-rich tasks.

\begin{figure*}[t]
    \centering
    \includegraphics[width=0.95\linewidth]{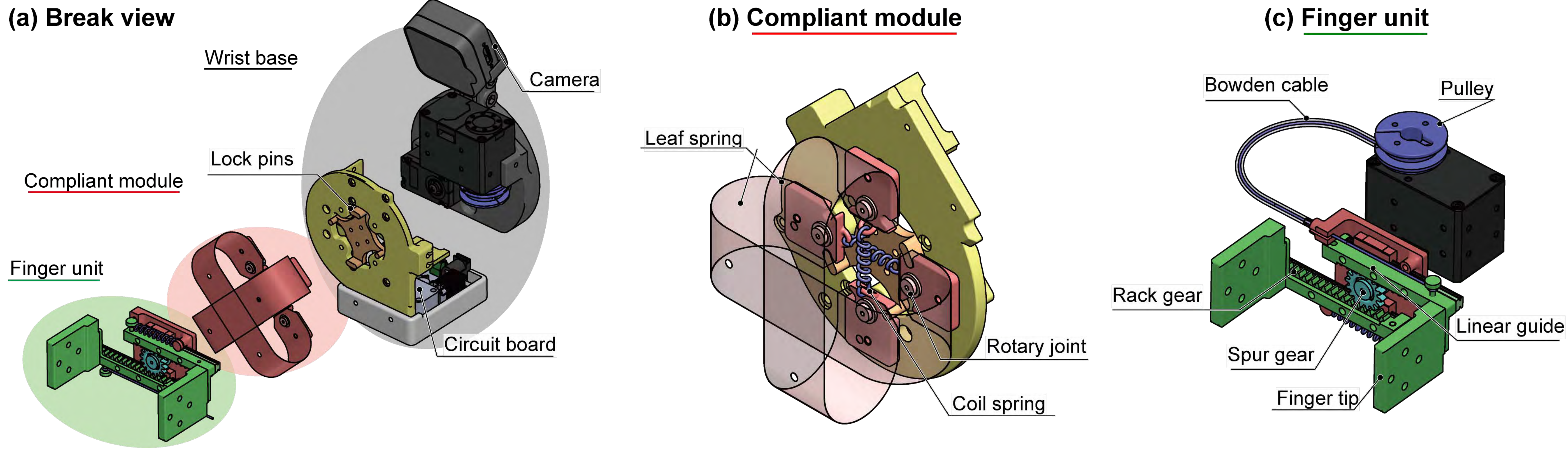}
    \caption{{\bf Mechanical design of the CLAW:} Orthogonal leaf springs and rotary joints provide 6-DoF compliance with anisotropic stiffness. A simple locking mechanism enables mode-switchable stiffness while preserving low weight and structural simplicity.}
    \label{fig:CAD}
\end{figure*}

\section{CLAW DESIGN CONCEPT}
The objective of this study is to develop a mechanism
for robust learning in contact-rich manipulation.
To this end, we propose a novel soft wrist mechanism, CLAW, which integrates a two-finger parallel gripper. The mechanical design of CLAW is presented in \cref{fig:CAD}. The overall system comprises three principal components: the finger unit, the compliant module, and the wrist base.
The compliant module constitutes the core of the mechanism, for which we introduce a convex tape–based leaf-spring structure.
This structure has the following features:
    \paragraph {\bf Large and multi-directional deformation} Leaf springs made of thin metallic tapes exhibit a wider deformation range and higher fatigue resistance compared to plastic flexure materials.
    \paragraph {\bf Anisotropic and adjustable stiffness} Thin metallic structures are highly compliant in bending and torsion, while remaining stiff in the shear direction. The stiffness is adjustable by a lock mechanism.
    \paragraph {\bf Lightweight construction} Metallic materials possess a high specific strength, and the thickness of the metallic tapes is less than 1 mm.
    \paragraph {\bf Low cost and simple implementation} By employing convex tapes as leaf springs, the compliant module inherently preserves a stable shape without the need for additional supporting components.

In the following subsections, we describe the deformation principle, provide theoretical analyses, and detail the remaining components of the system.

\subsection{Deformation Principle of Compliant Module}
\label{sec:soft_wrist}

\begin{figure}
    \centering
    \includegraphics[width=0.9\linewidth]{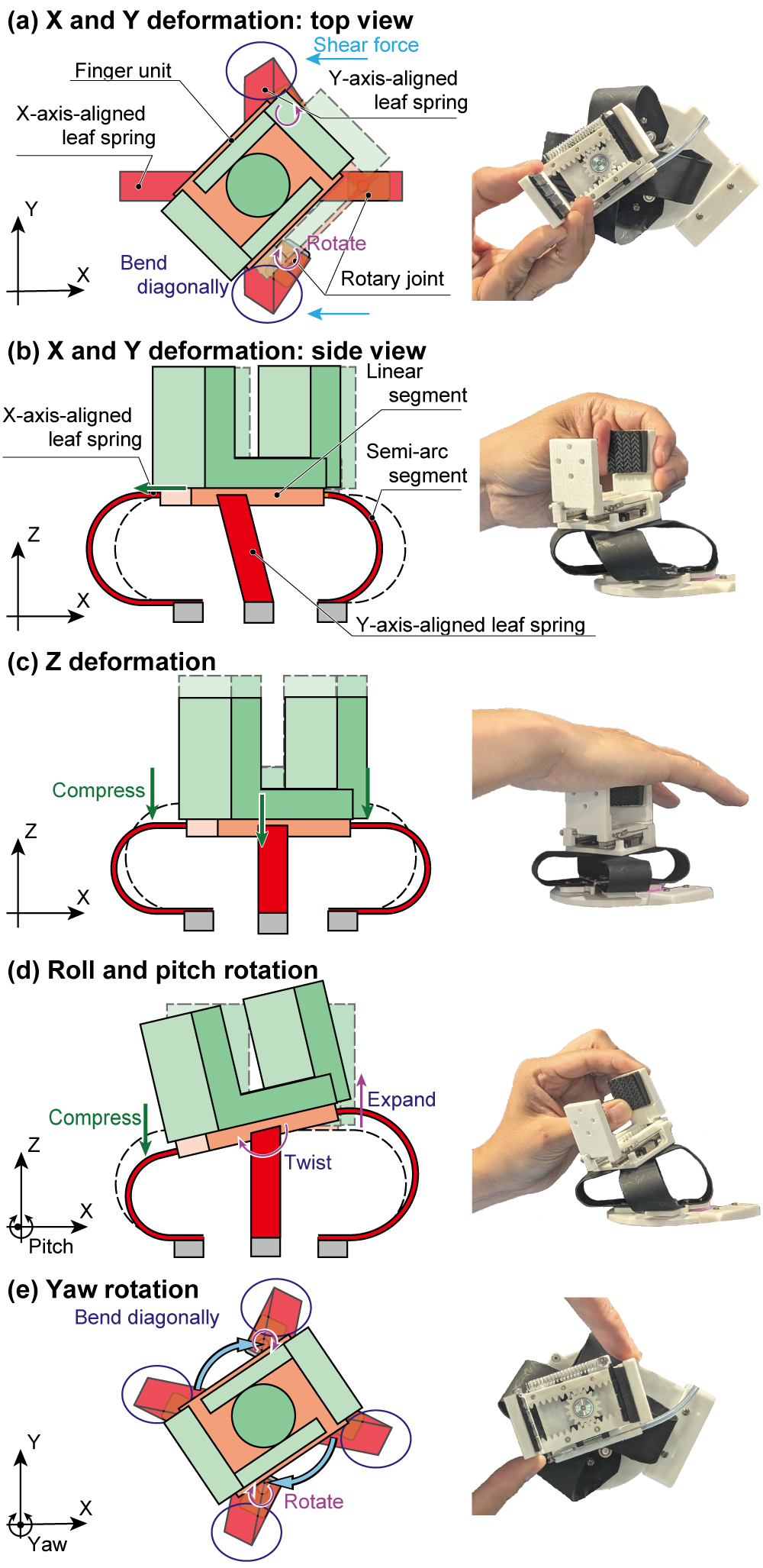}
    \caption{\textbf{Deformation characteristics of the CLAW compliant module:} Two orthogonal looped leaf springs with rotary joints and coil spring preload enable 6-DoF anisotropic compliance.}

    \label{fig:Deformation_of_leaf_spring}
\end{figure}

The compliant module contains two orthogonally arranged, looped leaf springs coupled to rotary joints pre-loaded by two coil springs, shown in~\cref{fig:Deformation_of_leaf_spring}~(a).  With only four compliant components, it provides 6-DoF and anisotropic compliance. The middle of the leaf springs is fixed to the finger unit, whereas both ends are attached to the rotary joints on the wrist base. Anisotropic stiffness arises from the combined bending and twisting of the leaf springs.
Furthermore, variable stiffness in X, Y, and yaw directions is obtained by selectively locking the rotary joints as shown in \cref{fig:CAD}~(b). CLAW has three lock modes: 1) Free mode releases all rotational joints, 2) Half-lock mode locks two joints in the X direction, and 3) Full-lock mode locks all four joints. 
The following explains the deformation mechanism for each degree of freedom.

\textbf{X and Y deformation} When an external force is applied along the
X- or Y-axis, the springs deform as illustrated in ~\cref{fig:Deformation_of_leaf_spring}~(b): the two bending (semi‑circular) sections translate, shifting the finger‑unit. When forces are applied in the shear direction to the leaf spring, displacement occurs by the rotary joints and the springs' bending and torsion, as shown in~\cref{fig:Deformation_of_leaf_spring} (a). If the rotary joints are locked while a force is applied (e.g., along the X-axis), the spring oriented along Y resists shear deformation because of its stiffness, thereby limiting displacement in the X direction.

\textbf{Z, roll, and pitch deformation} As shown in~\cref{fig:Deformation_of_leaf_spring}~(c) and (d), displacement along the Z-axis and rotation in roll or pitch occur due to changes in the radius of curvature of the bending parts (semi-arc segments) of the leaf springs. Also, in roll and pitch deformation, the leaf springs accommodate these movements through their twisting (center of~\cref{fig:Deformation_of_leaf_spring}~(d)).

\textbf{Yaw deformation} As shown in~\cref{fig:Deformation_of_leaf_spring}~(e), yaw deformation occurs when the rotary joints rotate in the same direction. This causes the leaf springs to undergo combined bending and torsional deformation, enabling the finger unit to rotate. When the rotary joints are locked, the shear stiffness of the leaf springs restricts yaw rotation.

\subsection{Deformation analysis of Compliant Module}
Expanding the compliant module's range of motion requires a corresponding increase in the looped leaf spring's initial lateral width. Below, we discuss the relationship between the size of CLAW’s loop structure and its range of motion in the XY plane.

\begin{figure}[t]
    \centering
    \includegraphics[width=\linewidth]{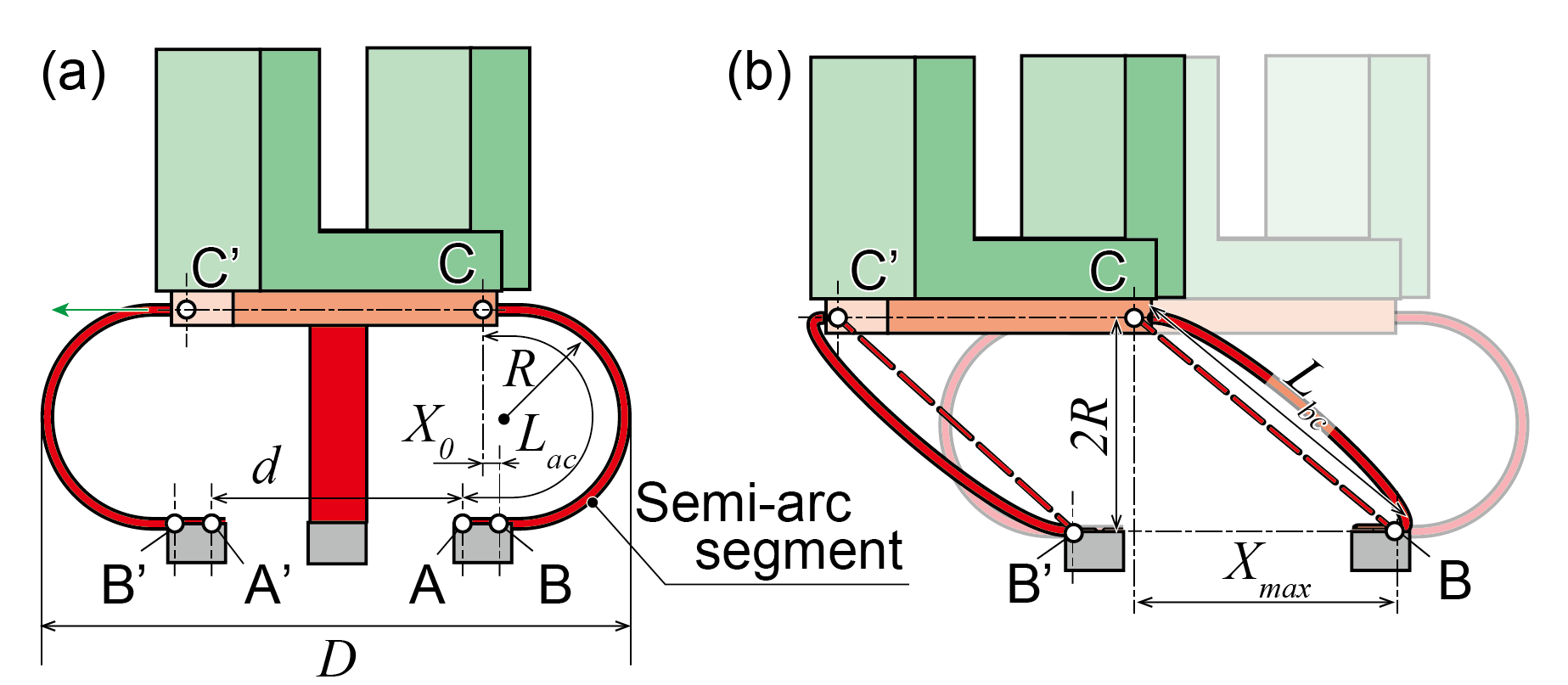}
    \caption{{\bf Length and range of motion of the leaf spring:} (a) Fixed section of the leaf spring and dimensions of each part, (b) Maximum deformation state in the XY direction. }
    \label{fig:deformation-analysis}
\end{figure}

\textbf{Definition of leaf‑spring lengths and loop width}
The deformations occur in the semi-arc segment of the leaf springs. 
Their lengths are defined as shown in~\cref{fig:deformation-analysis}. Points $A$ and $A'$ correspond to the axes of the rotational joints, and $C$ and $C'$ denote the clamped sections to the Finger unit. 
$L_{\alpha - \beta}$ represents a distance measured along the spring between two points $\alpha$ and $\beta$, and thus $L_{a - a'}$ is the total length of a leaf spring.
Given that the loop shape is approximated as a combination of semicircular arcs and straight segments, the lateral width of the loop $D$ is given by the following expression: 
\begin{equation}
    D = \frac{L_{a - a'} + d}{2} - \pi R + 2R,
\end{equation}
where $d$ is the inter‑axial distance between the $a$ and $a'$ rotational joints, determined by clearances with other parts.

\textbf{Range of motion in XY plane}
The maximum deformation in the XY plane is determined by the length of the leaf spring. When the gripper is displaced in the XY direction, the boundary between the semicircular segment and the straight segment first coincides with point $B$. As the displacement increases further, the radius of curvature of the semicircular arc gradually increases while deforming in the $X$ direction. For simplicity, it is assumed that when the gripper reaches its maximum displacement, the leaf spring becomes completely straight. Under this assumption, the maximum displacement of the gripper $X_{max}$ can be expressed as:
\begin{equation}
    X_{\max} = \sqrt{{L_{b-c}}^2 - 4R^2} - X_0.
\end{equation}
where $X_{0}$ is the initial position of $C$. 
Additionally, $L_{b-c}$ and $L_{a-a'}$ exhibit a linear relationship:
\begin{equation}
    L_{b-c} = \frac{L_{a-a'} - L_{c-c'}}{2} - L_{a-b}.
\end{equation}

In practical design, $D$ is first determined by size constraints, which in turn define $L_{a-a'}$.
We adopt a leaf spring with $R = 15$~mm and $L_{a-a'} = 180$~mm, yielding $D = 90$~mm, which corresponds to 1.2 times the UR5e's flange diameter.
Additionally, the system is operated such that the fingertip position error remains within $X_{max}$.
Deformation to this maximum value may lead to mechanical failure at the joints or other parts of the leaf spring. Therefore, in this study, the maximum allowable displacement is defined as 80\% of $X_{max}$, resulting in an allowable displacement of 38~mm.

\subsection{Convex-tape}
While a recent study proposes convex tapes for a gripper~\cite{he2025grasping}, we use them for the wrist compliant module. The tapes’ elastic restoring forces stabilize the finger unit’s pose while providing compliance in multiple directions. A convex tape is a metal strip with a uniform, arched cross‑section that is inherently stable in a straight configuration. When a convex tape is bent, the cross-section at the bending location flattens locally, allowing deformation only at that specific region.
Consequently, a loop formed from this material is not circular; instead, it has an oval‑slot shape composed of two semicircular arcs connected by straight segments. This geometry shortens the structure in the Z'-direction and lowers the moment transmitted to the springs by the finger unit’s weight when the end-effector is oriented horizontally.

\subsection{Lock Mechanism}

\begin{figure}[t]
    \centering
    \includegraphics[width=\linewidth]{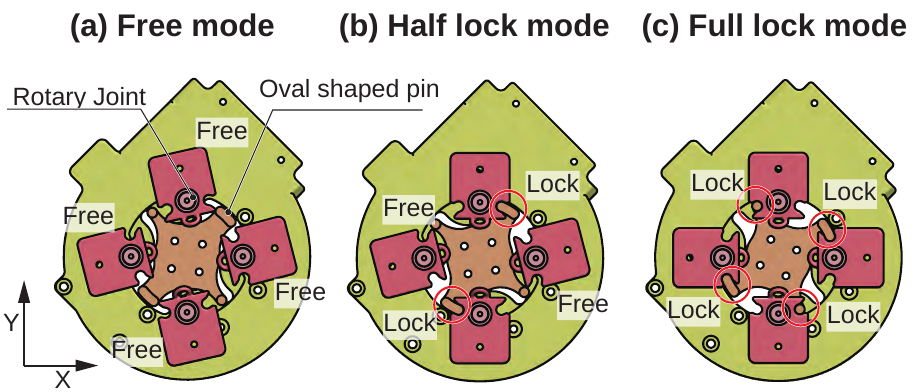}
    \caption{\textbf{Locking mechanism of the CLAW:} A single actuator moves four pins with different profiles on a carrier to engage grooves in joint links. This system enables three stiffness modes: (a) Free mode with all joints rotatable, (b) Half-lock mode restricting motion along the X-axis by locking two joints, and (c) Full-lock mode locking all joints to increase stiffness in X, Y, and yaw while retaining compliance in other directions.}
    \label{fig:lock}
\end{figure}

\cref{fig:lock} illustrates the proposed lock mechanism. Each joint link contains a groove, and locking is achieved by inserting pins into these grooves. All four pins are mounted on a single carrier and therefore move synchronously under a single actuator. Because the pins have different profiles, the joints that lock depend on the actuator’s direction of rotation, enabling three distinct stiffness modes.

\textbf{Free mode} (\cref{fig:lock}(a)) None of the pins engage their grooves, so all four joints remain fully rotatable.

\textbf{Half‑lock mode} (\cref{fig:lock}(b)) Only the slot‑shaped pins engage, locking two opposing joints; motion along the X-axis is restrained, while compliance in the Y-axis is preserved. This mode is advantageous when gravity acts along X, as it increases stiffness solely in the gravity direction.

\textbf{Full‑lock mode} (\cref{fig:lock}(c)) All pins engage, locking every joint. The compliant unit thus becomes stiff in X, Y, and yaw, whereas stiffness about
Z, roll, and pitch remains unchanged. This configuration suppresses vibration of the finger unit during free‑space motions while still permitting absorption of impact forces along Z during contact.

\subsection{Finger Unit and Wrist Base}
To prevent gravity-induced sagging of the springs, the finger unit is actuated via a remote Bowden-cable transmission. The wrist base houses two Dynamixel motors, including an XM430 for finger actuation and an XL330 for stiffness switching, together with the control electronics and an Intel RealSense D405 depth camera. 

\section{Experimental Evaluation}
To validate our system, we performed real robot experiments. Our goal is to answer the following questions: (1) Is CLAW more compliant in multiple directions (in~\cref{sec:deform_eval}), (2) Does the adjustable stiffness benefit in contact-rich manipulation (in~\cref{sec:benefit_var}), (3) Does CLAW complete contact-rich tasks more successfully than the other end-effectors (in~\cref{sec:benchmark_eval}), and (4) Is our system capable of executing real-world contact-rich manipulation tasks  (in~\cref{sec:varioustask_eval}).

To address these questions, first, we conducted deformation tests to characterize CLAW’s compliance by comparing different end-effectors. Second, we tested the adjustable stiffness effects in a door-opening task. Third, we evaluated its learning performance in a peg-in-hole benchmarking task against the end-effectors. Finally, we deployed CLAW on a variety of real-world manipulation tasks.

\subsection{Experimental Setting}
\label{sec:experiment-setting}
\textbf{Hardware} We used a collaborative robot arm (UR5e, Universal Robots A/S, Denmark) equipped with two cameras (RealSense D405, Intel Corporation, USA) and a desktop PC powered by a CPU AMD Ryzen Threadripper 7980X and a GPU NVIDIA RTX 6000 Ada.
In the deformation and benchmark evaluations, we compared CLAW with two baselines: a rigid gripper (2F-85, Robotiq, Canada) and a TPU-printed Fin Ray gripper, a common soft end-effector that is easy to fabricate. The policy was executed at 50 Hz.

\begin{figure}
    \centering
    \includegraphics[width=0.9\linewidth]{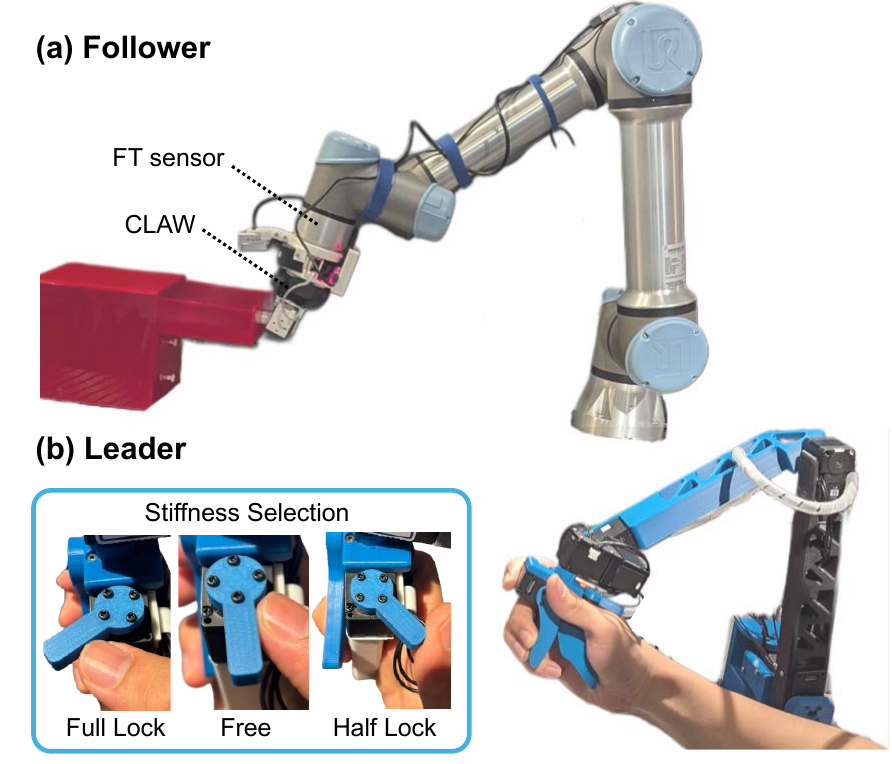}
    \caption{{\bf Teleoperation setup:} A haptic teleoperation leader device is employed to control a UR5e robot arm. On the leader device, user can select stiffness mode by switching a lever. FT sensor on the follower arm relay external force information to the leader, which is then mapped to joint torques to create haptic feedback.}
    \label{fig:teleop_setup}
\end{figure}

\begin{figure*}[t]
    \centering
    
    \includegraphics[width=1.0\linewidth]{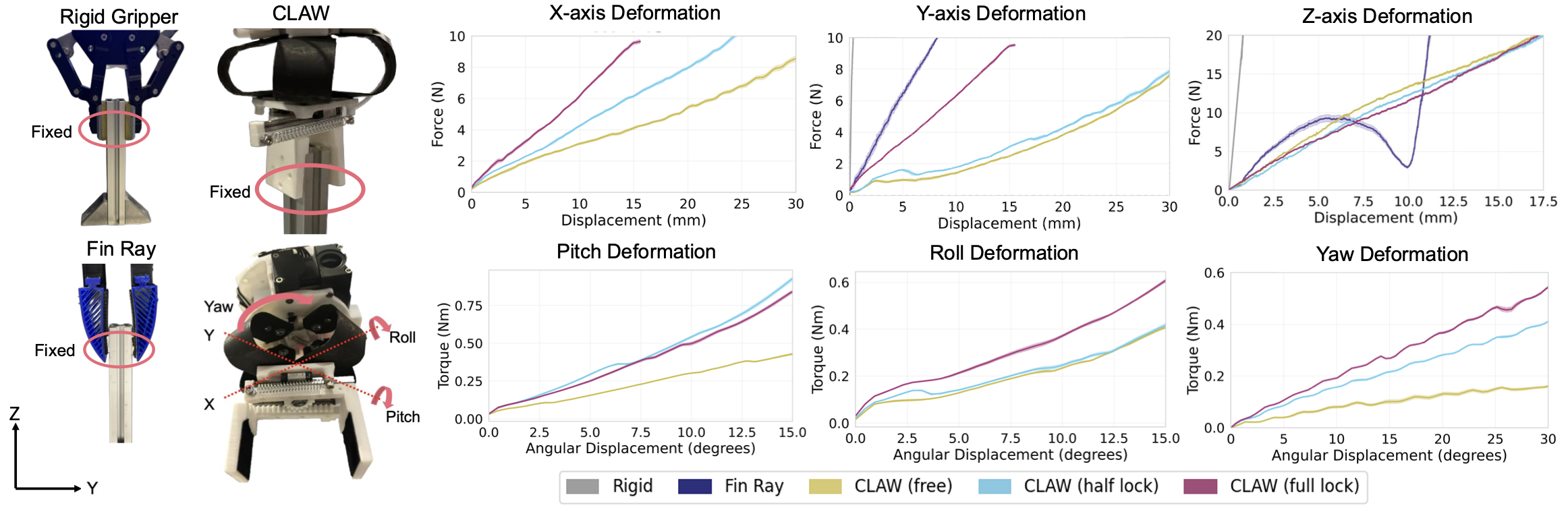}
\caption{\textbf{Deformation comparison of CLAW, Fin Ray, and a rigid gripper:}. Each end-effector’s tool center point was fixed to a vertical pole, as shown on the left side. For the Z-axis deformation on the Fin Ray, we placed the gripper on top of the vertical pole. Force readings were measured via the UR5e’s onboard FT sensor.}
    \label{fig:deformation-eval}
\end{figure*}

\textbf{Teleoperation system}
We developed a haptic feedback teleoperation system inspired by GELLO~\cite{wu2024gello} (leader robot) and its extension~\cite{sujit2025improving,liu2025factr}, which has an equivalent kinematic structure to the follower robot (UR5e), see \cref{fig:teleop_setup}.
The frame can be 3D-printed, and the servo motors are low-cost and can operate with current control.
While the prior study utilized the leader robot for a torque control-based robot~\cite{sujit2025improving, liu2025factr}, our system was designed for a position-control robot by
the two-channel teleoperation scheme from \cite{88057}: the leader streams joint positions to the follower, while the follower returns force feedback to the leader. 
We also added a lever switch for the user to easily change the stiffness of CLAW. 

\textbf{Imitation learning} We employed ACT~\cite{zhao2023learning}, a widely used imitation learning method known for its sample efficiency. We collected demonstrations using the teleoperation system.
ACT's inputs were images from wrist-mounted and fixed-view cameras, force-torque (FT) sensor readings, joint positions, and stiffness modes, and the outputs were action chunks: sequences of joint positions and stiffness modes.

\textbf{Control} \label{sec:control-strategy}
We employed a forward dynamics compliance controller~\cite{fdcc} to reduce oscillations and enable more stable teleoperation and policy execution during contact. The controller parameters were manually fine-tuned to minimize oscillations while preserving responsiveness for each gripper. The control loop ran at 500~Hz.

\subsection{Deformation Evaluation}
\label{sec:deform_eval}
We imposed incremental deformations by applying small translational (X, Y, Z) and angular (roll, pitch, yaw) motions and recorded the reaction wrench using the UR5e six-axis FT sensor. 
\cref{fig:deformation-eval} shows the deformation comparison.
Only the deformations of interest (Y- and Z-axis translations) were evaluated for the Fin Ray and the rigid gripper; therefore, the X-axis and rotational plots in~\cref{fig:deformation-eval} show CLAW only.
The Fin Ray is not compliant along the other axes, and the rigid gripper is stiff along all axes.

CLAW exhibits anisotropic stiffness. The Free mode is the softest, and the Full-lock mode is the stiffest. The Half-lock mode behaves similarly to Free along the Y-axis but is noticeably stiffer along the X-axis. For rotations, Half-lock is close to Full-lock in pitch but remains close to Free in roll. Along the Z-axis, the three CLAW modes have similar compliance. By contrast, the rigid gripper is highly stiff; Fin Ray's response is non-monotonic and unpredictable, stiffening initially, softening near 10 mm, and then stiffening again. In the Y-axis, the Fin Ray reaches about 10 N at 8 mm displacement, and the rigid gripper shows almost no compliance. Within CLAW, Full-lock requires roughly two times the force of Free at 15 mm in the Y-axis; in yaw, Full-lock delivers about three times the torque of Free at \(30^\circ\), whereas Half-lock delivers about two times the torque. 

Overall, CLAW enables compliant interaction in more directions than the baseline grippers, which are compliant only in a limited set of directions.
With these stiffness modes being selectable on demand, CLAW can be advantageous for robust and reliable contact-rich manipulation.

\begin{figure}
        \centering
        \includegraphics[width=1\linewidth]{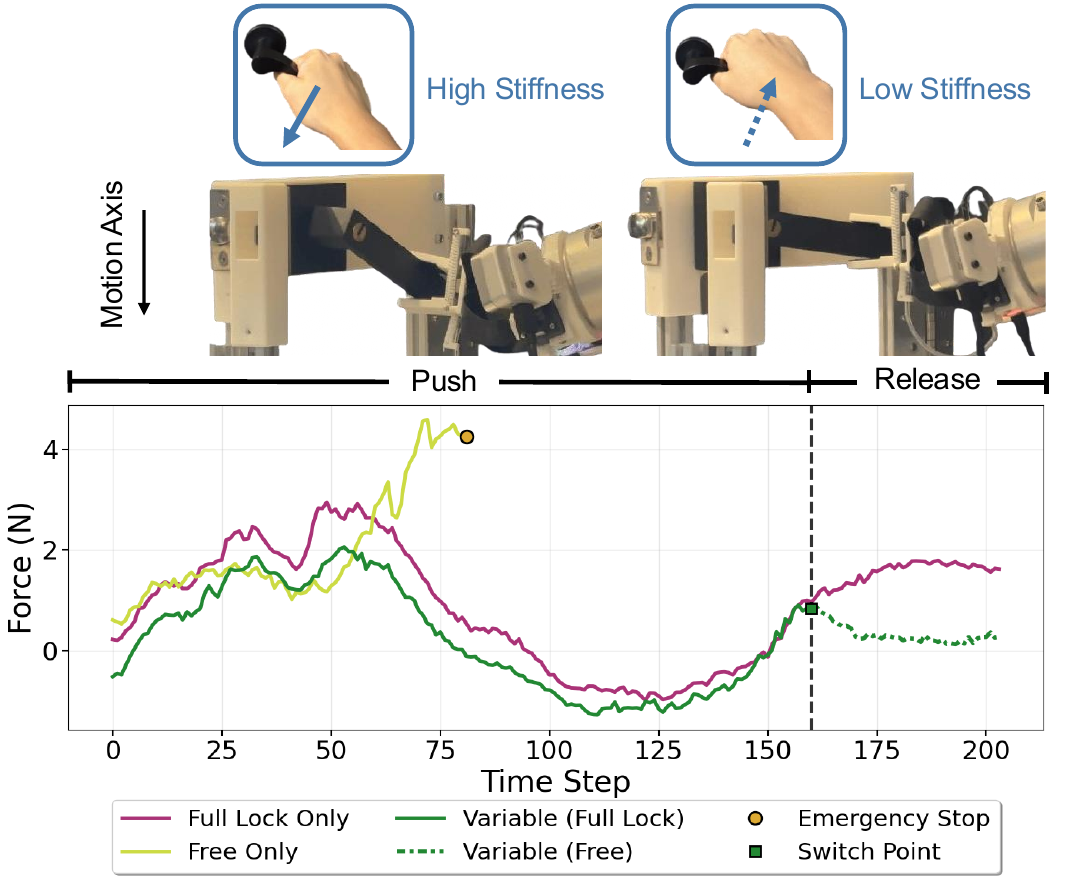}
        \caption{{\bf Door opening test:} Top figures show the motivation for variable stiffness in door opening as humans turn doorknobs with high wrist stiffness and release with low wrist stiffness. The bottom figure shows a comparison of the force trajectories of variable stiffness, free mode only, and full-lock only. Free mode triggered the emergency stop as the force threshold was exceeded in a different axis. }
        \label{fig:door-opening}
\end{figure}
\subsection{Benefit of Variable Stiffness}
\label{sec:benefit_var}

To validate the benefit of the variable stiffness mechanism, we examine whether switching stiffness mode mid-task provides a benefit to performance in a door-opening task.

In this experiment, we teleoperate CLAW using the system reported in \cref{sec:experiment-setting} to open a spring-loaded door handle. The door latch releases only after the handle is rotated by approximately 45°. As illustrated in \cref{fig:door-opening} (top), we follow a human-like strategy: a high-stiffness (Full-lock) mode during the push/rotation phase to transmit torque efficiently, then an immediate switch to low-stiffness (Free) mode once the latch releases. In Free mode, the wrist back-drives, so the door handle’s internal return spring sets the final pose without the robot resisting it. Then, we replayed the recorded episode in three stiffness conditions: (1) the recorded condition in which CLAW switches from Full-lock mode to Free mode, (2) only using Full-lock mode, and (3) only using Free mode. 

\cref{fig:door-opening} (bottom) shows the force profiles during the motion replays with the three conditions.
CLAW, with only Free mode, could not transmit enough force through the motion and triggered the emergency stop. Both Full-lock and variable stiffness modes were able to complete the task. By looking at the force trajectory along the primary motion axis, it can be observed that in Full-lock, elevated force persists even after the handle has rotated, indicating the wrist keeps resisting the handle’s return spring during the release phase. In contrast, variable stiffness switches to Free immediately at release, producing a sharp drop of force as it lets the handle’s internal spring set the terminal pose.

This stiffness switch is advantageous for the door opening. By handing control back to the environment after opening, the policy no longer needs to plan or time a precise “rotate-back” motion; Free mode's low stiffness passively accommodates the external spring. This reduces the need to actively exert forces (fewer protective stops).

\subsection{Learning Performance}
\label{sec:benchmark_eval}

\begin{figure}
    \centering
    \includegraphics[width=1\linewidth]{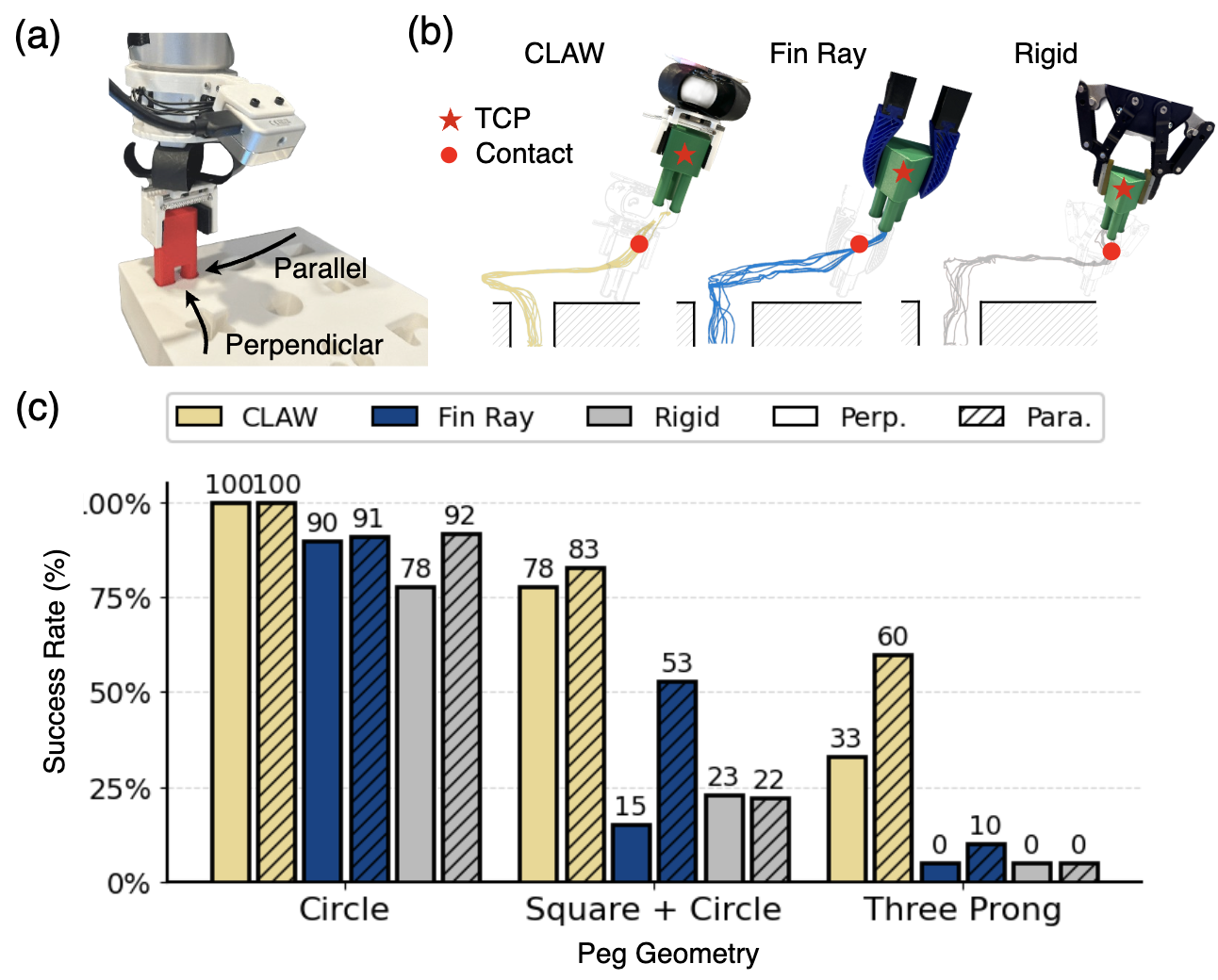}
    
    \caption{{\bf Learning performance benchmarking:} (a) Experimental setup with the FMB board and insertion peg. (b) Representative trajectory from the rollout insertion path of the three end-effectors. (c) Success rate comparison between the learning performance of CLAW, Fin Ray, and rigid gripper for different peg geometries and approach directions. }
    \label{fig:fmb-benchmark}
\end{figure}

To test whether CLAW has superior learning performance than off-the-shelf end-effectors, we conducted a benchmarking experiment using the assembly board from a Functional Manipulation Benchmark for Generalizable Robotic Learning (FMB)~\cite{doi:10.1177/02783649241276017}. 
We evaluated the three end-effectors on three peg-in-hole geometries representing the best-, median-, and worst-performing cases in the benchmark~\cite{doi:10.1177/02783649241276017}.
Since in different insertion paths the end effector may need to be compliant in different directions, we collected 25 human demonstrations from paths both parallel and perpendicular to the gripper palm (in \cref{fig:fmb-benchmark}(a)), yielding 50 demonstrations per task. We set a timeout of 60 seconds for inference.


By plotting multiple rollout trajectories (in \cref{fig:fmb-benchmark}(b)) from the benchmarking experiment, it can be observed that CLAW more effectively absorbs and suppresses post-contact oscillation. As a result, insertion paths are smooth and consistent across trials, demonstrating CLAW’s ability to maintain more precise, reliable motion in contact-rich scenarios.

The success rates (in \cref{fig:fmb-benchmark}(c)), were computed from policies trained on three random seeds, with each seed evaluated over 20 rollouts per approach direction (in total 120 trials). 
For the simpler circle peg, all grippers achieved relatively high success rates. 
For the square+circle peg, the Fin Ray gripper showed strongly divergent performance depending on approach direction. It achieved about four times higher success in the parallel direction, as its compliant axis helped with alignment and insertion along that path. The rigid gripper struggled in both directions, while CLAW maintained strong performance across both. For the three-prong peg, success rates dropped further due to the triangular layout (one peg at the top and two at the bottom). This configuration made the perpendicular approach particularly prone to collisions, resulting in a large discrepancy in approach directions. Despite this inherent geometric difficulty, CLAW still outperformed the baselines with an overall 76\% success rate across geometries, compared to 43\% and 36\% for the Fin Ray and rigid gripper, demonstrating robustness to approach directions and geometries. 

CLAW's main failure modes were slight misalignment or exceeding the time limit for each trial. Fin Ray often failed due to slippage caused by contact deformation. The rigid gripper frequently exerted excessive lateral (X–Y) forces that led to emergency stops. 

\subsection{Task Versatility}
\label{sec:varioustask_eval}
To demonstrate the versatility of our system, we evaluated CLAW on a range of challenging, delicate, and contact-rich tasks. For operations on vertical surfaces, such as horizontal peg-in-hole and drawer pulling, we engaged Full-lock mode to stiffen the end-effector when oriented horizontally, preventing sagging. Half-lock mode was chosen for dragging metal pegs, as its anisotropic stiffness allowed the gripper to transmit force along the motion axis while remaining laterally compliant to prevent jamming. For more complex tasks, including door opening and key insertion, stiffness modes were varied during the task. Finally, Free mode was used for the remaining tasks. We collected 30 demonstrations for single-stiffness tasks and 50 for variable-stiffness tasks. Success rates are reported from 20 rollout episodes. 

Results in \cref{tab:real-life} show the versatility of CLAW across a variety of real-world tasks by leveraging different stiffness modes. In peg-in-hole tasks with tight tolerances (approximately 50 ${\mu}$m), CLAW enables precise and compliant insertion without applying excessive force. Likewise, in glass-tube insertion, it ensures stable contact and minimizes the risk of damaging delicate objects. By actively varying its stiffness, CLAW adapts seamlessly to tasks such as key insertion and door opening. For key insertion, it first operates in Free mode to allow easier alignment, then switches to stiff mode to transmit sufficient torque for key rotation. For the door opening, as described in \cref{sec:benefit_var}, stiff mode is engaged initially to rotate the handle, after which CLAW transitions to Free mode to robustly release it. The supplementary material includes a video of these task executions.

\begin{table}[t]
\centering
\setlength{\tabcolsep}{10pt}
\renewcommand{\arraystretch}{1.3}
\caption{Task success rates for various real-life tasks.}
\label{tab:task_success_rates}
\footnotesize
\begin{tabular}{lcc}
\hline
\multicolumn{1}{c}{\textbf{Task}} &
\textbf{\begin{tabular}[c]{@{}c@{}}Success \\ Rate\end{tabular}} &
\textbf{\begin{tabular}[c]{@{}c@{}}Stiffness \\ Mode\end{tabular}} \\ \hline
Vertical peg-in-hole  & 0.90 & Free \\
Glass-tube insertion   & 0.85 & Free \\
Dish-cap placement    & 0.68 & Free \\
Drawing smiley face   & 0.47 & Free \\
Dragging metal peg    & 1.00 & Half-lock \\ 
Horizontal peg-in-hole  & 0.45 & Full-lock  \\
Drawer pulling        & 0.75 & Full-lock \\
Key insertion         & 1.00 & Free $\rightarrow$ Full-lock \\ 
Door opening          & 1.00 & Full-lock $\rightarrow$ Free \\ \hline
\end{tabular}
\label{tab:real-life}
\end{table}


\section{Conclusion}
\label{sec:conclusion}
This study introduced a novel soft wrist, CLAW, enabling robust learning of contact-rich manipulation tasks. Experiments showed that CLAW achieved higher success rates on benchmark tasks than widely used baseline grippers and successfully executed more demanding, contact-rich tasks.

{\bf Limitations and Future Work} Despite its promise, CLAW has several limitations.
First, it occasionally failed because the grasped object slipped. This issue could be mitigated by increasing fingertip friction or by using a higher-torque motor. Also, it may sag under the weight of excessively heavy objects.
Future work will apply CLAW to the other learning algorithms, such as reinforcement learning.

\bibliographystyle{IEEEtran}
\bibliography{main}

\end{document}